\documentclass[conference]{IEEEtran}
\IEEEoverridecommandlockouts
\usepackage{cite}
\usepackage{amsmath,amssymb,amsfonts}
\usepackage{algorithmic}
\usepackage{graphicx}
\usepackage{textcomp}
\usepackage{xcolor}
\usepackage{multirow}
\def\BibTeX{{\rm B\kern-.05em{\sc i\kern-.025em b}\kern-.08em
    T\kern-.1667em\lower.7ex\hbox{E}\kern-.125emX}}
\renewcommand{\thetable}{\arabic{table}}
\begin{document}

\title{Incorporating the Knowledge Distillation to Improve the EfficientNet Transfer Learning Capability}

\author{\IEEEauthorblockN{Tjeng Wawan Cenggoro}
\IEEEauthorblockA{\textit{Computer Science Department, School of Computer Science} \\
\textit{Bioinformatics and Data Science Research Center} \\
\textit{Bina Nusantara University}\\
Jakarta, Indonesia, 11480\\
wcenggoro@binus.edu}
}

\maketitle

\begin{abstract}
Due to the Deep Learning requirement for a large training dataset, Transfer Learning has become a central method in the field of Computer Vision, which heavily used Deep Learning. Since the adoption of transfer learning in the field, the performance of the models in computer vision is significantly upgraded. The common transfer learning practice in computer vision research is to use state-of-the-art architectures in the ImageNet dataset as the backbone network. The best performing architecture in ImageNet is currently held by EfficientNet, which is a set of architectures that were progressively grown from a baseline architecture.

By observing the EfficientNet development approach, it is natural to hypothesize that the smaller EfficientNet architecture is capable to contain the knowledge of the larger architecture. This hypothesis comes from the fact that the larger EfficientNet architecture was grown from the smaller architecture. Therefore, in this study, we experimented if it is beneficial to transfer the knowledge from the larger architecture to the smaller architecture of the EfficientNet. To achieve this goal, we proposed a transfer learning method that uses Knowledge Distillation as the knowledge transfer mechanism from the larger to smaller architecture. We found that the proposed method is able to upgrade the performance of each of the EfficientNet architecture. Several smaller architectures are even able to outperform the larger architecture that was trained using the standard transfer learning method. 
\end{abstract}


\section{Introduction}
In the last decade, Deep Learning has become the method of preference for most of Computer Vision problems. However, it has a noticeable weakness that demands a large dataset for training \cite{pardamean2019using}. Without any solution to this weakness, Deep Learning can be difficult to be adopted in real-world problem, because of the cost to collect the large dataset. Fortunately, it is possible to overcome this weakness with a strategy called Transfer Learning. In this strategy, a deep learning model can be trained optimally with a medium-sized dataset. The optimal performance can be achieved because the model has been pretrained with the available large dataset.

\begin{figure}[t]
\centerline{\includegraphics[width=0.9\hsize]{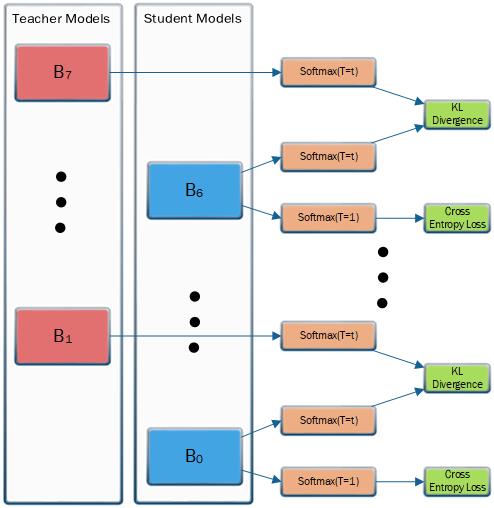}}
\caption{The illustration of the proposed transfer learning method using the Knowledge Distillation for the EfficientNet architectures.}
\label{fig:proposed}
\end{figure}

Specifically in image classification, a model named as EfficientNet \cite{tan2019efficientnet} emerged to be the current state-of-the-art in the largest and most popular image classification dataset, ImageNet \cite{deng2009imagenet, russakovsky2015imagenet}. Apart from its performance in ImageNet, the EfficientNet also demonstrated an excellent capability for transfer learning to other image classification datasets. In this paper, we report our experiment to improve the transfer learning capability of the EfficientNet. The proposed method is to utilize the use of the Knowledge Distillation technique \cite{hinton2015distilling} for exploiting the hierarchical architecture of the EfficientNet model variation. This proposed method is illustrated in Fig. \ref{fig:proposed} and explained in details in section \ref{method}.

\section{Related Work}
The current transfer learning practice to train deep learning models was shaped by Girshick et al. \cite{girshick2014rich} study in developing an object detection model. In the study, they used a Convolutional Neural Network (CNN) \cite{lecun1989backpropagation} that was trained to classify an image with the ImageNet dataset. To transform the CNN to be an object detector, they added extra modules on the top of the last CNN layer that can be trained to detect objects. This approach happened to be state-of-the-art in Pascal VOC \cite{everingham2010pascal}, the most popular object detection at that time. Given the size of Pascal VOC that is much smaller than ImageNet, the transfer learning approach proposed by Girshick et al. was proved to solve the deep learning issue that requires a large training dataset.

Since the success of R-CNN, their transfer learning approach became a norm not only in object detection  \cite{ren2015faster,redmon2016you,lin2017focal,tan2019efficientdet} but also in other computer vision problems such as instance segmentation \cite{chen2019hybrid,he2017mask}, action recognition \cite{carreira2017quo}, and crowd counting \cite{cenggoro2019feature,yan2019perspective,ma2019bayesian,liu2019crowd}. The transfer learning approach used in these studies follows the same procedure as the R-CNN, which is to use the ImageNet pretrained network as a backbone and then add auxiliary modules that are specific to the new tasks.

In practice, the pretrained network used as the backbone in many studies were the winners of ImageNet Large Scale Visual Recognition Competition \cite{krizhevsky2012imagenet,simonyan2015very,szegedy2015going,he2016deep} or the state-of-the-art model in ImageNet at the time \cite{huang2017densely,tan2019efficientnet}. Because the original study of the pretrained network usually did not include the transfer learning analysis, other studies did the analysis \cite{yosinski2014transferable,kornblith2019better}. However, the study of the current ImageNet state-of-the-art (EfficientNet) \cite{tan2019efficientnet} included the transfer learning analysis of the model as well. 

\section{Methodology}
\label{method}
For a more understandable mathematical explanation, in this paper, we used the notation $B_i$, where $i \in \{0,1,2,3,4,5,6,7\}$ to represent the EfficientNet architecture variation. Thus, for the baseline EfficientNet architecture, it is represented as $B_0$ instead of the original notation B0 in this paper.

In the EfficientNet study, the compound model scaling is used to grow $B_i$ to $B_{i+1}$ sequentially from $i=0$ to $i=7$. Having observed this approach, we hypothesize that $B_i$ and $B_{i+1}$ have an architectural relationship. Therefore, if we can transfer the knowledge from the trained $B_{i+1}$ model to $B_i$ model, it should be possible to have a model with the size of $B_i$ but with the performance of $B_{i+1}$, which always has a better performance as reported by Tan and Le \cite{tan2019efficientnet}.

To enable the knowledge transfer from $B_{i+1}$ to $B_i$, we employ the Knowledge Distillation (KD) technique \cite{hinton2015distilling}. It is currently the only method that is suitable for this case. This technique allows knowledge transfer from a cumbersome model that acts as a teacher to a smaller model that acts as a student. In the KD, the student model needs to minimize the cross entropy loss of its class probabilities $p_s(x)$ given input $x$ in regards to the teacher model probabilities $p_s(x)$. This cross entropy loss can be formulated using the Kullback-Leibler Divergence $D_{KL}(p_t|p_s)$ as follows:

\begin{multline}
H(p_t,p_s) = \sum_{x \in X} p_t(x) log(p_t(x)) + D_{KL}(p_t|p_s) \\
= \sum_{x \in X} p_t(x) log(p_t(x)) + \sum_{x \in X} p_t(x) log(\frac{p_s(x)}{p_t(x)})
\label{eq:cel}
\end{multline}

\noindent in which the probability $p$ for a class $c$ is defined as follows:

\begin{equation}
p(x) = \frac{exp(z_c(x)/T)}{\sum_{c \in C} exp(z_c(x)/T)}
\label{eq:prob}
\end{equation}

\begin{figure}[htbp]
\centerline{\includegraphics[width=0.9\hsize]{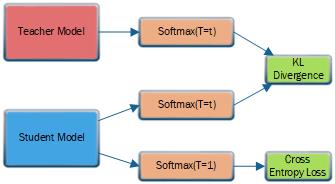}}
\caption{The illustration of the knowledge distillation process.}
\label{fig:kd}
\end{figure}

\noindent where $z_c(x)$ is a model that produces logits for class $c$, $T$ is the temperature parameter to control the softness of probabilities for the KD. The probabilities need to be softened to ease the learning process. When $T=1$,  (\ref{eq:prob}) is collapsed to the standard softmax. It is important to notice that, in the KD, only the student model is trained while the teacher model is frozen. Consequently, the class probabilities $p_t(x)$ becomes a constant in (\ref{eq:cel}), hence only the KL Divergence term needs to be minimized. The student model is also trained with the true label as the target by using the standard cross entropy loss. The overall process of the KD is illustrated in Fig. \ref{fig:kd}.

\begin{figure}[b]
\centerline{\includegraphics[width=\hsize]{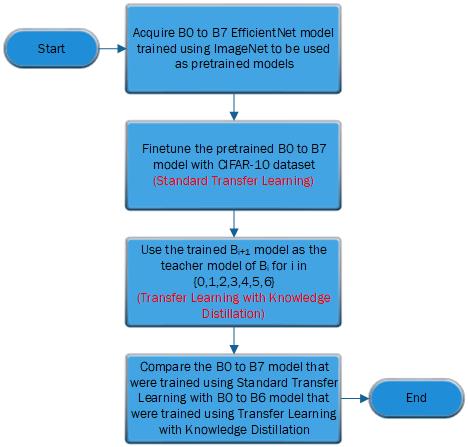}}
\caption{The research scenario.}
\label{fig:steps}
\end{figure}

As illustrated in Fig. \ref{fig:proposed}, we first trained the EfficientNet architecture $B_0$ to $B_7$ with the target dataset for transfer learning. In this study, we used the CIFAR-10 dataset \cite{krizhevsky2009learning} as the target transfer learning dataset. This dataset is often used as the benchmark for transfer learning \cite{kornblith2019better}. Afterward, we trained the EfficientNet architecture $B_0$ to $B_6$ by using the KD with the $B_1$ to $B_7$ as the teacher model respectively. These steps are summarized in Fig. \ref{fig:steps}.

\begin{figure}[t]
     \centering
     \includegraphics[width=0.5\textwidth]{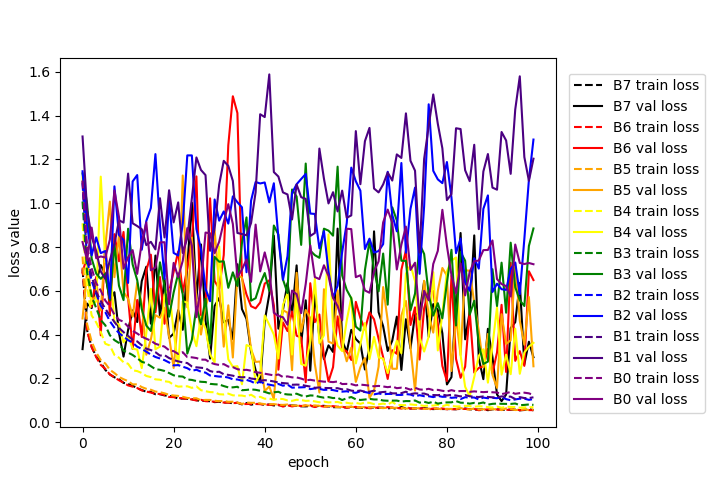}
    \caption{Teacher models training performance plot, smoothed by the weight equals to 0.3.}
    \label{fig:teachers-plot-0.3}
\end{figure}

\begin{figure}[t]
     \centering
     \includegraphics[width=0.5\textwidth]{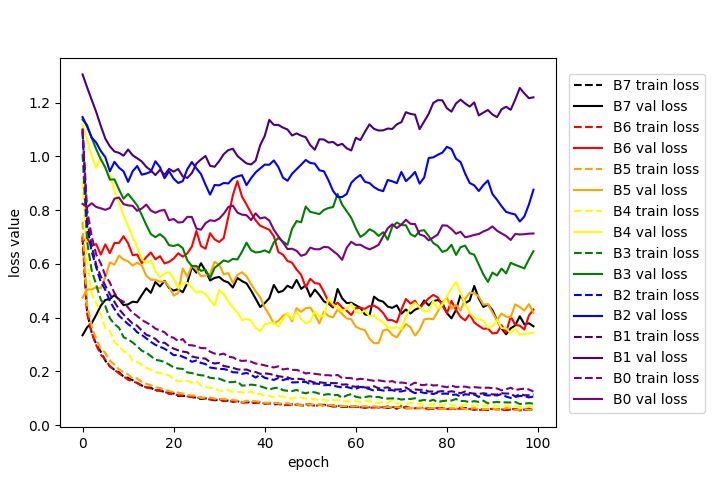}
    \caption{Teacher models training performance plot, smoothed by the weight equals to 0.9.}
    \label{fig:teachers-plot-0.9}
\end{figure}

The training process of all models, including the teachers and the students, used the Adam optimization method \cite{kingma2015adam} with the default setting, the L2 regularization rate of $10^{-5}$, and the batch size of 32. The temperature used in the KD for training the students is set to 2. Following the image resolution scaling in the original EfficientNet study, we also used a scaled resolution with the same ratio as in the original study. Starting from the CIFAR10 original resolution of $32\times32$ for the $B_0$, the resolution is scaled to $34\times34$, $38\times38$, $44\times44$, $54\times54$, $66\times66$, $76\times76$, and $86\times86$ for $B_1$ to $B_7$.  

\section{Results and Discussions}
\begin{figure}[t]
     \centering
     \includegraphics[width=0.5\textwidth]{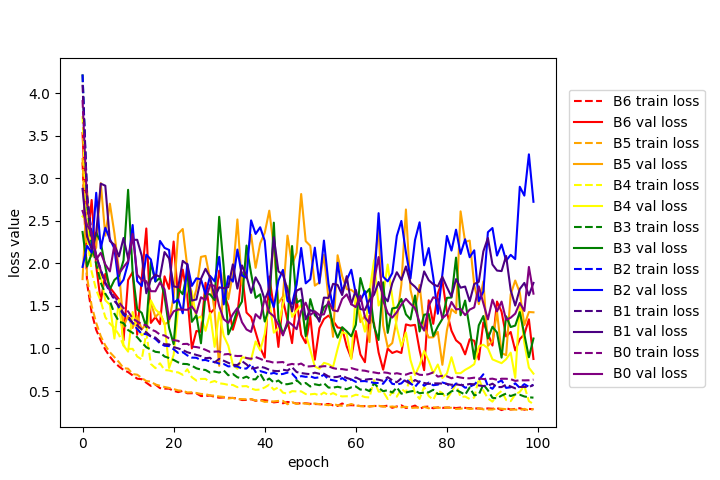}
    \caption{Student models training performance plot, smoothed by the weight equals to 0.3.}
    \label{fig:students-plot-0.3}
\end{figure}

\begin{figure}[t]
     \centering
     \includegraphics[width=0.5\textwidth]{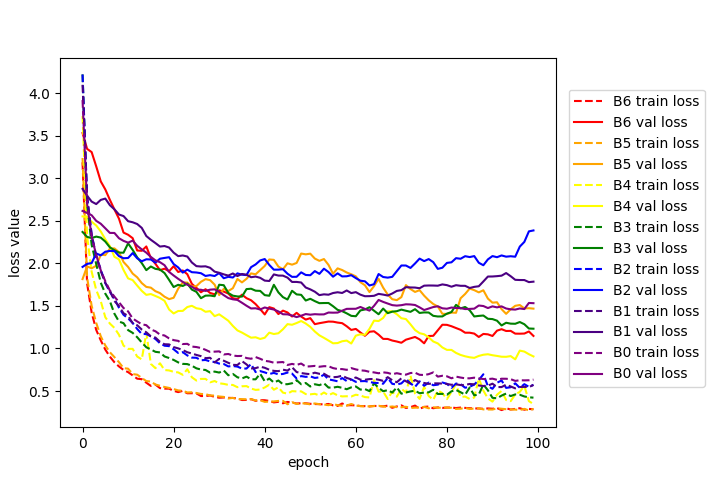}
    \caption{Student models training performance plot, smoothed by the weight equals to 0.9.}
    \label{fig:students-plot-0.9}
\end{figure}

To see whether the models suffered overfitting, we plotted the training and validation loss of the teacher Fig. \ref{fig:teachers-plot-0.3} and \ref{fig:teachers-plot-0.9} while the students plots are depicted in Fig. \ref{fig:students-plot-0.3} and \ref{fig:students-plot-0.9}.
For a better view, we smoothed the validation curve with the exponential moving average with the weight equals 0.3 for Fig. \ref{fig:teachers-plot-0.3} and \ref{fig:students-plot-0.3} while the weight is equal to 0.9 for Fig. \ref{fig:teachers-plot-0.9} and \ref{fig:students-plot-0.9}. From the plots, we can infer that the student models suffer less overfitting compared to the teacher models. This result can be explained by the regularization effect of the KD, as described by Hinton et al. \cite{hinton2015distilling}.

In Table \ref{tab:comparison}, we reported the accuracy comparison of our proposed transfer learning method versus the standard transfer learning method for the EfficientNet. With the same architecture, our proposed method consistently outperforms its standard method counterpart. 

\begin{table}[htbp]
\caption{The performance comparison of the standard transfer learning versus the transfer learning with knowledge Distillation.}
\begin{center}
\begin{tabular}{|c|c|c|c|}
\hline
\textbf{Base} & \textbf{Standard} & \textbf{Transfer Learning with}\\
\textbf{Model} & \textbf{Transfer Learning} & \textbf{the Knowledge Distillation}\\
\hline
$B_7$ & 93.61\% & - \\
$B_6$ & 93.23\% & \textbf{93.71\%} \\
$B_5$ & 92.55\% & \textbf{93.28\%} \\
$B_4$ & 92.11\% & \textbf{92.68\%} \\
$B_3$ & 90.79\% & \textbf{91.50\% }\\
$B_2$ & 89.44\% & \textbf{89.58\%} \\
$B_1$ & 87.69\% & \textbf{88.55\%} \\
$B_0$ & 87.35\% & \textbf{87.82\%} \\
\hline
\multicolumn{3}{l}{The accuracy in bold indicates a better performance}
\end{tabular}
\label{tab:comparison}
\end{center}
\end{table}

Surprisingly, the student models trained with our proposed method appear to be able to outperform its teacher model. In Table \ref{tab:teacher-student}, we rearrange Table \ref{tab:comparison} so that the teacher accuracy and the student accuracy is next to each other. We can see that $B_6$, $B_5$, $B_4$, and $B_0$ outperform their teacher model. From this result, we can conclude that the architecture of these four models is able to contain the information embedded in their teacher model. Thus, further training improves the accuracy of these student models.

\begin{table}[htbp]
\caption{The comparison of the teacher and student accuracy.}
\begin{center}
\begin{tabular}{|c|c|c|c|}
\hline
\textbf{Teacher} & \textbf{Student} & \textbf{Teacher} &  \textbf{Student}\\
\textbf{Model} & \textbf{Model} & \textbf{Accuracy} & \textbf{Accuracy} \\
\hline
$B_7$ & $B_6$ & 93.61\% & \textbf{93.71\%} \\
$B_6$ & $B_5$ & 93.23\% & \textbf{93.28\%} \\
$B_5$ & $B_4$ & 92.55\% & \textbf{92.68\%} \\
$B_4$ & $B_3$ & \textbf{92.11}\% & 91.50\% \\
$B_3$ & $B_2$ & \textbf{90.79}\% & 89.58\% \\
$B_2$ & $B_1$ & \textbf{89.44}\% & 88.55\% \\
$B_1$ & $B_0$ & 87.69\% & \textbf{87.82\%} \\
\hline
\multicolumn{4}{l}{The accuracy in bold indicates a better performance}
\end{tabular}
\label{tab:teacher-student}
\end{center}
\end{table}

\section{Conclusion}
\cite{semeniuta2016recurrent}
The experiment results in this study demonstrate that the proposed transfer learning method that utilizes Knowledge Distillation is beneficial to improve the transfer learning capability of the EfficientNet architectures. The proposed method is capable to produce a more accurate model with the same EfficientNet architecture. Moreover, the EfficientNet $B_6$, $B_5$, $B_4$, and $B_0$ that was trained with the proposed method are able to outperform its teacher model that has a larger architecture.

The future research direction following this study can explore the use of the KD in a more flexible scheme, that might allow the transfer from $B_7$ to $B_0$ for instance. Another promising direction is to use the KD to transfer learning from the EfficientNet models to other CNN models. 

\section*{Acknowledgement}
The experiment in this research was executed using the NVIDIA Tesla P100 supported by NVIDIA - BINUS AI R\&D Center. 

\bibliographystyle{./IEEEtran}
\bibliography{./IEEEabrv,./IEEEexample}

\end{document}